\documentclass[11pt, a4paper]{article}
\usepackage{rotating}
\usepackage[utf8]{inputenc}
\usepackage[T1]{fontenc}
\usepackage[english]{babel}
\usepackage[margin=2.5cm]{geometry} 
\usepackage{lmodern} 
\usepackage{natbib}
\usepackage{multirow}
\usepackage{subcaption}
\usepackage{graphicx}
\usepackage{booktabs}
\usepackage{amsmath, amssymb} 
\usepackage{authblk}

\usepackage[colorlinks=true,allcolors=blue]{hyperref}

\newcommand{\orcid}[1]{\href{https://orcid.org/#1}{\,(ORCID)}}

\title{\bfseries HideAndSeg: an AI-based tool with automated prompting for octopus segmentation in natural habitats}

\author[1]{Alan de Aguiar\thanks{\href{mailto:aln.deaguiar@gmail.com}{aln.deaguiar@gmail.com}}}

\author[1]{Michaella Pereira Andrade\thanks{\href{mailto:michapereirandrade@gmail.com}{michapereirandrade@gmail.com}}}

\author[1]{Charles Morphy D. Santos\thanks{\href{mailto:charlesmorphy@gmail.com}{charlesmorphy@gmail.com}}}

\author[1]{João~Paulo~Gois\thanks{\href{mailto:joao.gois@ufabc.edu.br}{joao.gois@ufabc.edu.br}}}

\affil[1]{Universidade Federal do ABC (UFABC), Santo André, SP, Brazil}

\date{} 

\begin{document}

\maketitle

\begin{abstract}
Analyzing octopuses in their natural habitats is challenging due to their camouflage capability, rapid changes in skin texture and color, non-rigid body deformations, and frequent occlusions, all of which are compounded by variable underwater lighting and turbidity. Addressing the lack of large-scale annotated datasets, this paper introduces HideAndSeg, a novel, minimally supervised AI-based tool for segmenting videos of octopuses. It establishes a quantitative baseline for this task. HideAndSeg integrates SAM2 with a custom-trained YOLOv11 object detector. First, the user provides point coordinates to generate the initial segmentation masks with SAM2. These masks serve as training data for the YOLO model. After that, our approach fully automates the pipeline by providing a bounding box prompt to SAM2, eliminating the need for further manual intervention. We introduce two unsupervised metrics – temporal consistency ($DICE_t$) and new component count ($NC_t$) – to quantitatively evaluate segmentation quality and guide mask refinement in the absence of ground-truth data, i.e., real-world information that serves to train, validate, and test AI models. Results show that HideAndSeg achieves satisfactory performance, reducing segmentation noise compared to the manually prompted approach. Our method can re-identify and segment the octopus even after periods of complete occlusion in natural environments, a scenario in which the manually prompted model fails. By reducing the need for manual analysis in real-world scenarios, this work provides a practical tool that paves the way for more efficient behavioral studies of wild cephalopods.
\end{abstract}

\maketitle

\section{Introduction}
To study wildlife, biologists often must manually analyze massive amounts of video footage, a process complicated by data collected over long periods with varying equipment, climate, geographical, and environmental conditions~\citep{Andrade2023,Lalgudi2025}. The challenge amplifies in complex marine environments, particularly for studying cephalopods, whose camouflage skills, swift changes in skin color and texture, and highly deformable bodies present unique analytical hurdles~\citep{Ikeda2021,Schnell2021}. 

Cephalopods exhibit the most rapid and varied skin color changes in the animal kingdom~\cite{Ikeda2021}. These bodily changes are referred to as body patterns and are based on the alteration of chromatophores, texture, posture, and locomotion~\citep{Messenger2001,Ikeda2021}.

Given the idiosyncrasies of cephalopods, automating key steps of cephalopod analysis in natural habitats is fundamental for advancing studies in animal ethology, welfare, and conservation. Initially, a solution to the task of reliable video segmentation is required. 

\begin{figure}[htbp]
    \centering
    \includegraphics[width=1\linewidth]{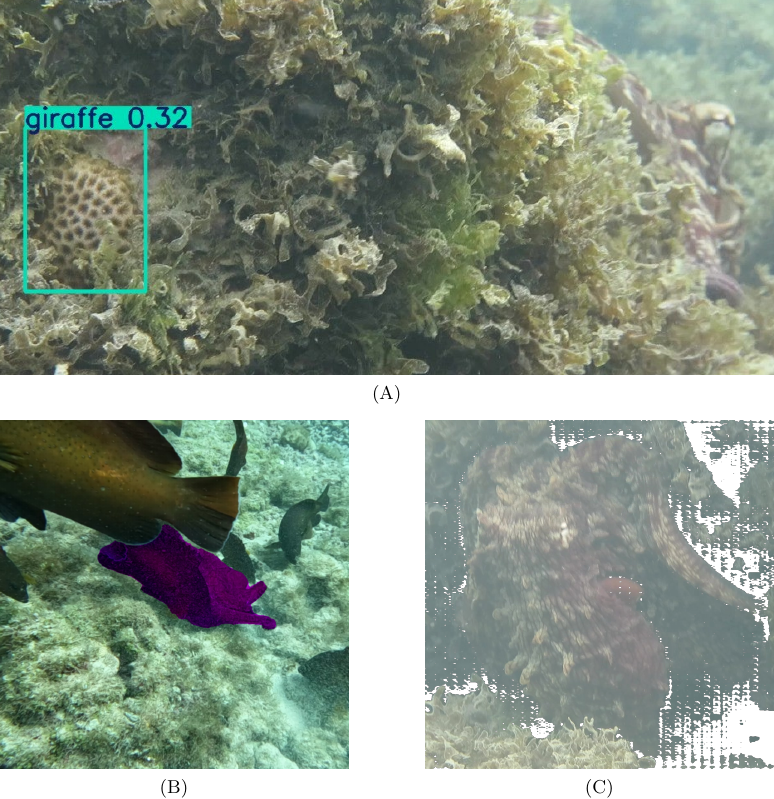}
    \caption{Failure cases for models when applied to octopus videos. (A) YOLO incorrectly labels the coral reef as a “giraffe” and fails to detect the octopus on the right side of the frame; (B) In SAM2, a fish crossing in front of the octopus causes the segmentation mask (a purple tint) to leak into the fish; (C) Also in SAM2, camouflage and water conditions cause the surrounding environment to leak into the segmentation mask.}

    \label{fig:1}
\end{figure}

In recent years, Artificial Intelligence (AI) systems like \emph{You Only Look Once} (YOLO) and segmentation models like the \emph{Segment Anything Model} (SAM) and its video-based successor, SAM2, have been adapted for wildlife detection~\citep{Chen2024,Roy2023,Ravi2024,Lalgudi2025}. Their ability to generate high-quality masks with minimal prompting makes them promising for marine biology, especially given the often inconsistent quality of their imagery data. However, these models can be brittle when applied directly to unconstrained videos of octopuses in natural environments. YOLO models struggle with the specific visual challenges of the underwater domain (Fig.~\ref{fig:1}-A). Likewise, SAM2 can track an octopus across several frames but may fail under challenging conditions such as camouflage, occlusions, or poor visibility (Fig.~\ref{fig:1}). 

These generalist object-detection and image segmentation AI-based models are not explicitly trained to handle the common issues of non-stationary appearance, non-rigid deformation, and environmental degradation that occur in natural environments. Specifically concerning cephalopods, another critical barrier is the absence of large-scale, publicly available, annotated video datasets for octopus segmentation, which prevents standard supervised fine-tuning and evaluation. 

Processing videos of octopuses recorded in natural environments is a core requirement for advancing behavioral research. Minimally invasive video recordings capture key behavioral states not previously reported in captive observations~\citep{Andrade2023}. Nevertheless, underwater videos are challenging to interpret due to the complex optical properties of aquatic environments, including color distortion, low contrast, blurring, noise, scattering, inconsistent lighting, and turbidity~\citep{Vijayalakshmi2024}. These factors degrade image quality and hinder accurate object detection and segmentation. To address these issues, deep learning and image processing techniques have been proposed, including convolutional neural network (CNN) and generative adversarial network (GAN) based image enhancement models, image formation models that account for underwater light propagation, generation of synthetic data to supplement limited real-world datasets, and domain adaptation via transfer learning~\citep{Vijayalakshmi2024}.

This study addresses the complexities of automated segmentation applied to octopuses observed within their natural, uncontrolled habitats. Overcoming the obstacles inherent in these tasks requires considering factors like: (1) Dynamic appearance: octopuses rapidly change skin color, pattern, and texture~\citep{Ikeda2021,Shook2024}. For a computer vision model, the target’s visual features are never stable, violating the assumption of appearance consistency that many tracking and segmentation algorithms rely on; (2) Articulated body: lacking a skeletal structure, an octopus can deform and contort its body into a vast range of shapes~\citep{Flash2023}. Tracking such a non-rigid object is more difficult than monitoring organisms with predictable skeletons; (3) Degraded underwater visuals: the underwater environment introduces visual distortions that degrade video quality~\citep{Vijayalakshmi2024}. These include turbidity, which obscures object boundaries; depth-induced color filtering that washes out key visual information; highly variable illumination, which creates shadows and reflections; and (4) the presence of other organisms: natural habitats are uncontrollable environments where other life forms, such as fish and seaweed, can obscure the octopus in video recordings and hamper automated tracking. 

We introduce HideAndSeg, a novel AI-based methodology for segmenting octopus videos in unconstrained underwater environments, built on the SAM2 and YOLO architectures and designed explicitly for segmenting octopuses in videos recorded in natural habitats. HideAndSeg works even without readily available ground truth data, i.e., with no objective standard of correctness, which makes it possible to both train and evaluate the performance of the video-segmentation model. To address the "no ground truth" challenge, we propose a workflow that minimizes manual intervention by leveraging unsupervised tracking and segmentation metrics for quantitative evaluation. Furthermore, the study identifies the limitations of current state-of-the-art models through comprehensive analysis, providing a clear roadmap for future research. This work distinguishes itself from prior studies by offering an image segmentation pipeline tailored to octopus videos in natural settings, which significantly reduces manual labeling efforts. It introduces unsupervised metrics to assess the quality of the generated segmentations.

\section{Material and Methods}
\subsection{Subjects, study area, and field data collection}
\emph{Octopus insularis} is a medium-sized benthic octopus that inhabits shallow waters from Brazil to the Caribbean, including oceanic islands~\citep{OBrien2021,Cortes2025}. Our observations and filming of juvenile O. insularis were conducted off the northeastern coast of Brazil, in three locations with rich ecosystems: Atol das Rocas Biological Reserve, Búzios Beach, and Fernando de Noronha Marine National Park. These environments, composed of rocks, coral reefs, and algae, enable us to study the behavioral patterns of octopuses in shallow, clear waters. We used a GoPro 10 and a Canon G16 camera for filming, following the octopuses as they foraged on the substrate. We maintained a distance of approximately two meters from the observed individuals, prioritizing non-interference with their behavior, as the research was minimally invasive and non-lethal. The second author of the work collected the data in seven expeditions from 2022 to 2024 within the project \emph{Investigating sentience and emotional states in wild octopuses}\footnote{\url{https://www.wildanimalinitiative.org/blog/grantee-octopus}}.

\subsection{YOLO and SAM}

YOLO~\citep{Redmon2015}  is an object detection model that uses a single convolutional neural network (CNN) to simultaneously predict multiple bounding boxes and their corresponding class probabilities in a single pass over the image. YOLOv11 extends beyond object detection to include pose estimation and instance segmentation. It also achieves an improved mean average precision (mAP) while using 22\% fewer parameters than YOLOv8 in the COCO dataset~\citep{Khanam2024}. 

Researchers have successfully used models from the YOLO family in wildlife identification tasks~\citep{Thomas2025}. The innovations introduced by this architecture have led to a reduction in inference time, which, combined with high accuracy, enables the use of these models in real-time scenarios. This results in a better trade-off between performance and speed when compared to other CNN-based object detection architectures~\citep{Thomas2025}. However, there are limitations regarding model generalization, as they may not perform well for species and environmental conditions that were not present in the training dataset, thus requiring properly labeled data~\citep{Thomas2025}. This requirement limits the use of pre-trained models in the context of octopus identification. 

Additionally, the complexity of images captured in underwater environments presents challenges, including distortions caused by light refraction in water and turbulence~\citep{Yu2025} (Yu et al., 2025). For such cases, it is possible to extend the architecture’s capability, as seen in TMAE-YOLO~\citep{Yu2025}, which introduces the TMAE (Temporally Multi-scale Attention Enhancement) and TD-AFPN (Temporal Decoupled Attention Feature Pyramid Network) modules to improve the detection of mud crabs underwater. TMAE leverages reference frames to integrate temporal information through a cross-frame attention mechanism, enabling the model to enhance feature representations by selectively focusing on informative regions across multiple frames, thus recovering details lost due to occlusion or poor visibility. TD-AFPN improves the feature extraction process, particularly for small and medium-scale objects, by decoupling spatial and temporal feature aggregation. The use of these modules resulted in improved performance in mud crab detection underwater, achieving an AP50 of 84.3\%, which represents an improvement over YOLOv8n's performance of 80.7\%. However, TMAE relies on reference frames for feature enhancement. In cases of rapid target motion or sudden appearance changes—such as abrupt viewpoint shifts or significant deformation—the cross-frame attention mechanism may struggle to align features accurately, thereby reducing its effectiveness. This challenge is crucial in the context of octopus identification, where camouflage and dynamic body movement are essential characteristics. 

Combining object segmentation with detection methods like YOLO~\citep{Redmon2015,Zheng2024} involves three steps: using video object segmentation to focus on the target organism and reduce background noise; retraining recognition networks with these segmented images to enhance individual identification; and developing a system capable of real-time detection, segmentation, and recognition. 

SAM is a general-purpose image segmentation framework that operates across various domains without requiring task-specific training or massive annotated datasets~\citep{Kirillov2023}. Given a simple input prompt, such as a bounding box or a point indicating the object of interest, SAM can generate high-quality segmentation masks, even for objects or image types not encountered during training. This flexibility makes SAM particularly useful in settings like ecological monitoring, where labeled data is limited. SAM2~\citep{Ravi2024} extends the original framework to support video segmentation, accepting point, box, or mask prompts on individual frames. It propagates segmentation across time, employing a lightweight, promptable mask decoder that combines image embeddings with user-provided prompts to produce segmentation masks frame by frame. The user can iteratively refine the prompts to enhance accuracy, making SAM2 a practical tool for image and video segmentation. 

Given the enormous volume of data and the difficulty of manually segmenting octopuses due to their complex body shapes, ability to camouflage, and environmental conditions, creating a fully annotated dataset for training specialized models is expensive. In this context, a generalist model such as SAM2 becomes valuable. 

Despite the generalization capability of SAM2, manual prompting remains a necessary step, which can be costly when processing large volumes of video data. To address this limitation, the FLAIR method~\citep{Lalgudi2025} was developed as an alternative, leveraging the CLIP (Contrastive Language–Image Pretraining) model, which learns joint representations between natural language and images. By constructing a specialized textual prompt, we used CLIP to identify which of the segmentation masks generated by SAM2 corresponded to the target object—specifically, sharks. The masks identified by CLIP were then used as input prompts to SAM2, enabling the tracking of the object throughout the entire video sequence. A significant advantage of this approach is that it does not require any annotated data; only the textual prompt used for mask classification needs to be adapted. However, the quality of the segmentation produced is entirely dependent on the performance of the CLIP model, which may struggle to identify specific species that are underrepresented in its pretraining dataset. To validate the method, in addition to frame-by-frame manual annotations, FLAIR was compared against object detection models~\citep{Lalgudi2025} (YOLOv8, DETR~\citep{Carion2020}) used for frame annotation, as well as a Human-in-the-Loop approach, in which a human operator provided manual prompts whenever segmentation was lost. The detection-based methods performed poorly on the test set, which negatively impacted the resulting segmentation quality—particularly because object detection outputs were used to prompt every frame, thereby propagating any noise introduced by the detector. In contrast, the Human-in-the-Loop strategy exhibited performance closely aligned with that of FLAIR, suggesting that targeted interventions at points of segmentation failure—potentially automated by a more robust object detector and mask quality metrics—may be sufficient to achieve high-quality results.

\begin{figure}
    \centering
    \includegraphics[width=1\linewidth]{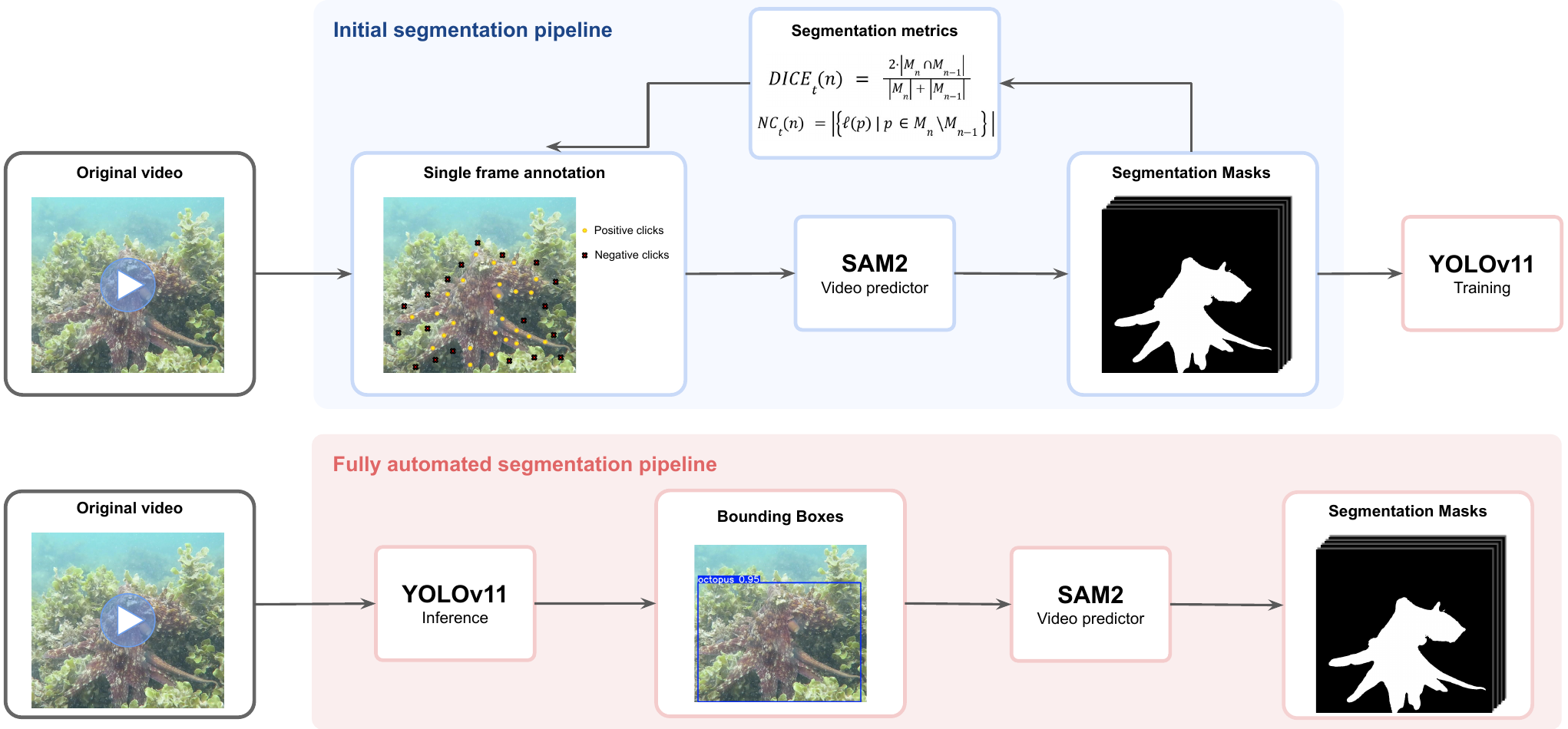}
    \caption{HideAndSeg pipeline. The input video is first processed through frame extraction. The first clear frame is manually annotated to provide an initial prompt for SAM2, which then generates segmentation masks that are evaluated using our proposed unsupervised metrics. For additional manual annotation, one can select the frame with the lowest metric score. Once the process is complete, the resulting masks are used to train a YOLO-based object segmentation model that ultimately replaces manual prompt annotation for SAM2, resulting in a fully automated segmentation process.}
    \label{fig:2}
\end{figure}

HideAndSeg relies on using SAM2 to generate segmentation masks from manual annotations on a small number of frames, and then leveraging these results to train a YOLO object detection model, thereby eliminating the need for human intervention (Fig.~\ref{fig:2}). To ensure the results are accurate and reliable, we compute unsupervised segmentation metrics throughout the pipeline. These metrics guide the refinement of the initial prompts used with SAM2 and assess the quality of the final segmentations.

\subsection{Video segmentation}
The initial dataset consisted of 148 octopus videos of varying lengths and frame rates, totaling 564,755 frames. We discarded all the initial frame sequences in which the octopus was not visible, resulting in empty segmentation masks, and then applied the segmentation process to the remaining 366,514 frames. We adopted the small and large variants of the SAM2 model, containing 46 million and 224 million parameters, respectively. For both models, we used the video predictor module provided by the SAM2 library, which propagates segmentations across the entire video based on annotation prompts. These prompts can be associated with any frame in the sequence. We enabled asynchronous frame loading to prevent memory overflow during processing. 

\subsection{Unsupervised evaluation metrics}

Standard video segmentation evaluation relies on comparing model outputs with ground-truth annotations, often using metrics such as the DICE coefficient. As our dataset is unlabeled, we employ two unsupervised metrics to assess different facets of segmentation quality from the raw model output. 

In high-frame-rate videos, the change in an object’s position and shape between consecutive frames is expected to be minimal. Leveraging this assumption, we use the $DICE_t$ coefficient to measure the temporal consistency of the segmentation by comparing the mask of the current frame, $M_n$, with the mask of the previous frame $M_{n-1}$ . Formally, the metric is: 
\begin{equation}
DICE_t(n) = \frac{2\left|M_n\right|\cap \left|M_{n-1}\right|}{\left|M_n\right| + \left|M_{n-1}\right|}.    
\end{equation}

A higher $DICE_t$ score indicates smooth and consistent segmentation, while a lower score suggests abrupt changes, potentially due to tracking loss or erratic mask generation. It is worth mentioning that to differentiate the unsupervised metric from the standard usage, we will use the notation $DICE_t$ to refer to the computation of the metric between the segmentation of two consecutive frames, and $DICE$ to denote the comparison of the segmentation with the ground truth.

A qualitative analysis of the segmentation results revealed that when the model fails to accurately identify the octopus, the output mask often becomes noisy and fragmented, characterized by numerous small, disconnected components (speckling), particularly at the edges (Fig.~\ref{fig:1}-C). While some disconnected components are expected due to natural occlusions from vegetation or rocks, a sudden increase in their number signals a drop in segmentation quality. 

To quantify this phenomenon, we employ a metric for measuring the deterioration of segmentation quality by counting the number of new connected components that appear in the current frame’s mask relative to the previous one. Let $\ell(p)$ be the label of the connected component to which the pixel p belongs. The metric reads: 

\begin{equation}
NC_t(n) =\left|\ell(p)~|~p\in M_n \setminus M_{n-1}\right|.
\end{equation}

To evaluate video segmentation stability, we determined the number of connected components $NC_t$. These components were identified using a full connectivity (8-connected) algorithm, as implemented in the scikit-image package~\citep{VanderWalt2014}. A low $NC_t$ value indicates a stable segmentation, while a high value points to the emergence of noise and potential tracking failure. 

\subsection{Manual frame annotation}
We tested three manual annotation strategies to assess their impact on segmentation performance. In the first run, a manual annotation was performed on the initial frame, supplying the coordinates corresponding to the octopus in the image. In the second run, negative annotations were also included, i.e., coordinates that should not be considered part of the target object. Then, $NC_t$ metric was used to select an additional annotation frame. For each video, the frame that exhibited the most significant increase in the metric relative to the previous frame was selected. The goal was to identify the exact moment when the segmentation became more unstable and provide additional information to help the model sustain its performance. We use both positive and negative annotations for this additional frame. Although SAM2 does not require a large number of annotated points to perform segmentation, challenges such as partial occlusion and camouflage can lead to fragmented masks with multiple disconnected components. To mitigate this, the selected coordinates were distributed as uniformly as possible along the contour of the octopus, rather than using a fixed number of keypoints, ensuring more complete and coherent segmentation results.

Finally, we performed manual segmentation on the initial frame of all videos to enable evaluation of the method using traditional segmentation metrics and to ensure the effectiveness of the proposed unsupervised segmentation metrics, which directly depend on the quality of the segmentation in the first frame.

\subsection{Object Detection}
While SAM2 significantly reduces the manual effort required for video segmentation, the annotation and validation process remains time-consuming when working with large-scale video datasets. We trained a YOLO-based model specialized in octopus detection to eliminate human intervention. Given the complexity and scale of our dataset, we selected YOLOv11-l, a model with 25.3 million parameters, pre-trained on the COCO dataset. 

For training, we used the results from the segmentation experiment that achieved the lowest NCt metric. From the segmentation masks, we extracted bounding boxes around the octopus in each frame to serve as ground truth. Frames with empty segmentation masks were excluded, resulting in a reduction of the total number of frames from 366,514 to 305,291.

We divided the remaining frames into training (212,924 frames), validation (56,288 frames), and test (36,079 frames) subsets. To prevent data leakage, we ensured that no video in the test set appeared in the training or validation sets. We evaluate the model performance on the test set using Precision, Recall, and Mean Average Precision (mAP) at two Intersection over Union (IoU) thresholds: 0.5 (mAP@0.5) and 0.95 (mAP@95). 

\subsection{Combining segmentation and detection}
\begin{figure}[htbp]
    \renewcommand{\thesubfigure}{\roman{subfigure}} 
    \centering
    \textbf{(A) SAM2 output} \\
    \begin{subfigure}{0.32\linewidth}
        \includegraphics[width=\linewidth]{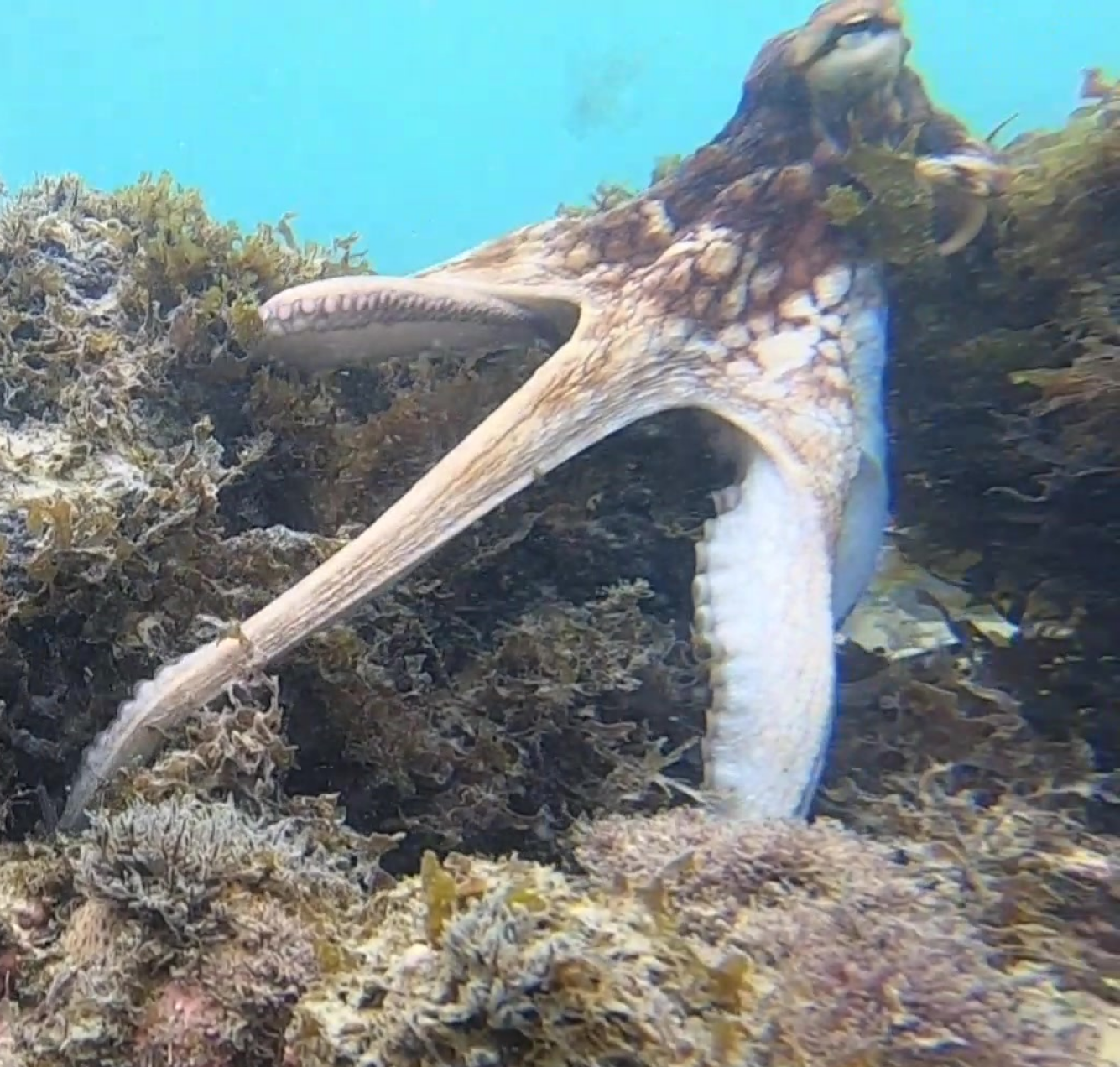}
        \caption{Frame $t$} 
        \label{fig:sam1}
    \end{subfigure}\hfill 
    \begin{subfigure}{0.32\linewidth}
        \includegraphics[width=\linewidth]{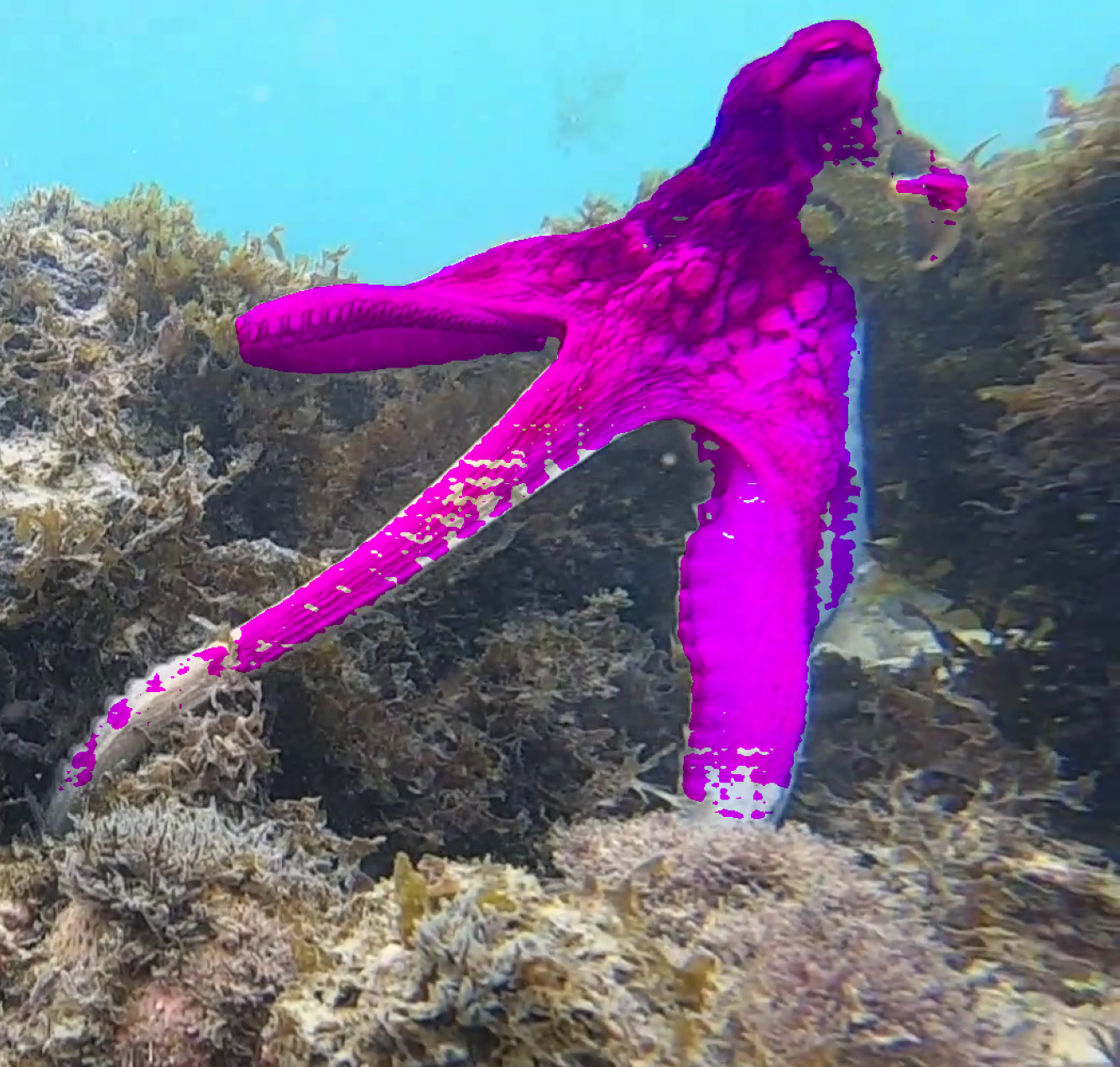}
        \caption{Frame $t+1$}
        \label{fig:sam2}
    \end{subfigure}\hfill
    \begin{subfigure}{0.32\linewidth}
        \includegraphics[width=\linewidth]{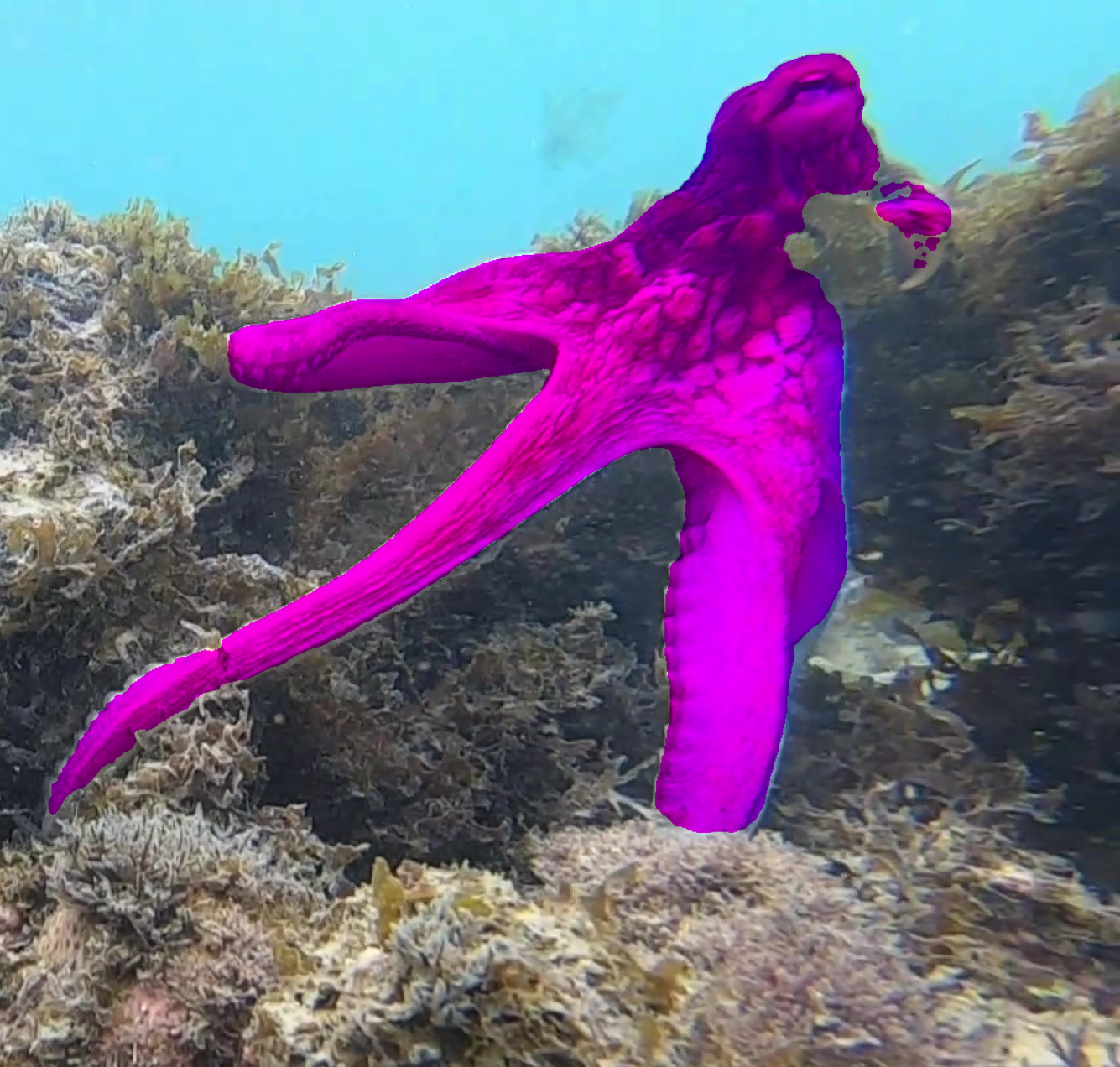}
        \caption{Frame $t+2$}
        \label{fig:sam3}
    \end{subfigure}    
    \vspace{0.5cm} 
    \setcounter{subfigure}{0} 
    \textbf{\\(B) YOLO output} \\
    \begin{subfigure}{0.32\linewidth}
        \includegraphics[width=\linewidth]{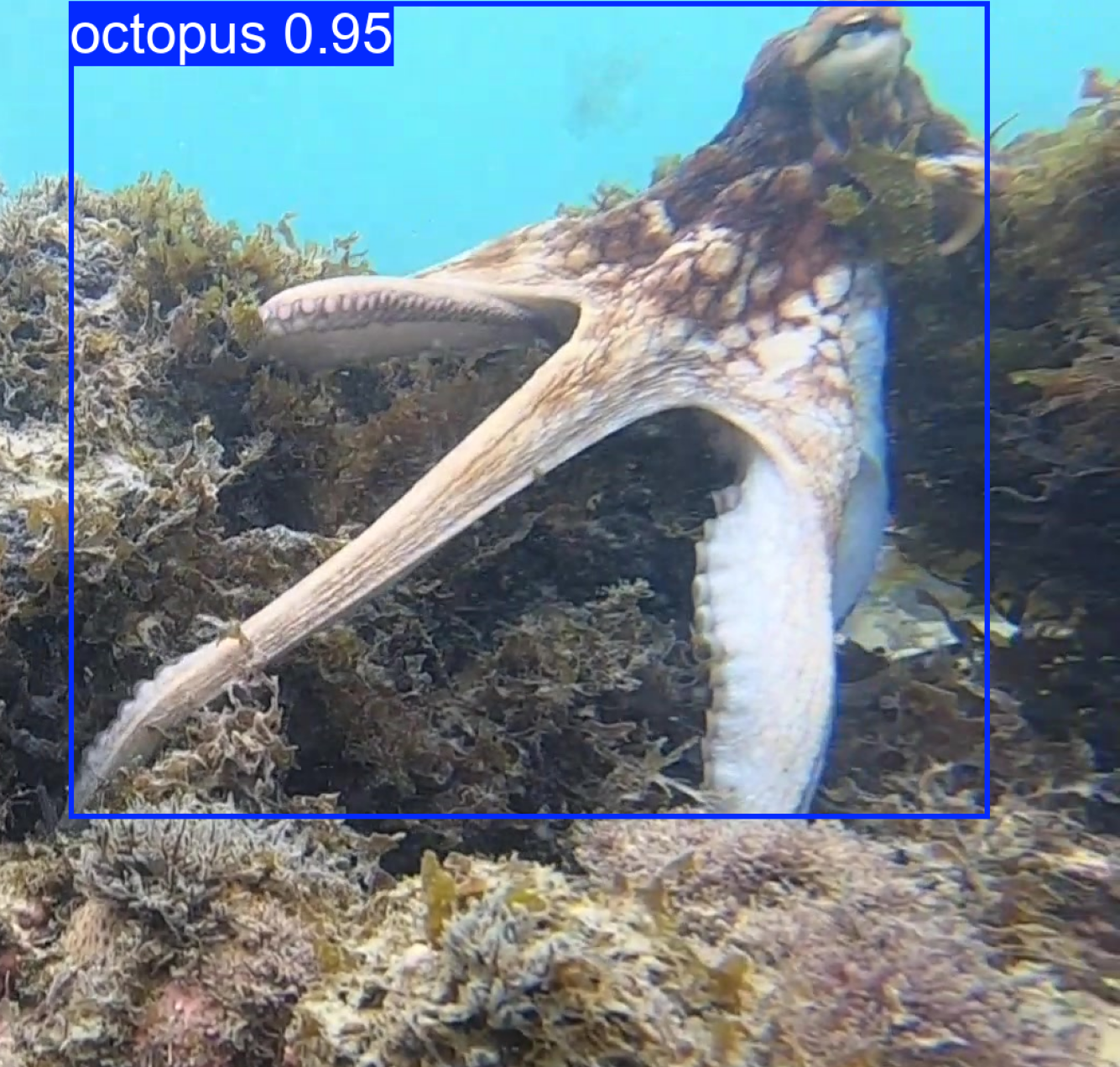}
        \caption{Frame $t$}
        \label{fig:yolo1}
    \end{subfigure}\hfill
    \begin{subfigure}{0.32\linewidth}
        \includegraphics[width=\linewidth]{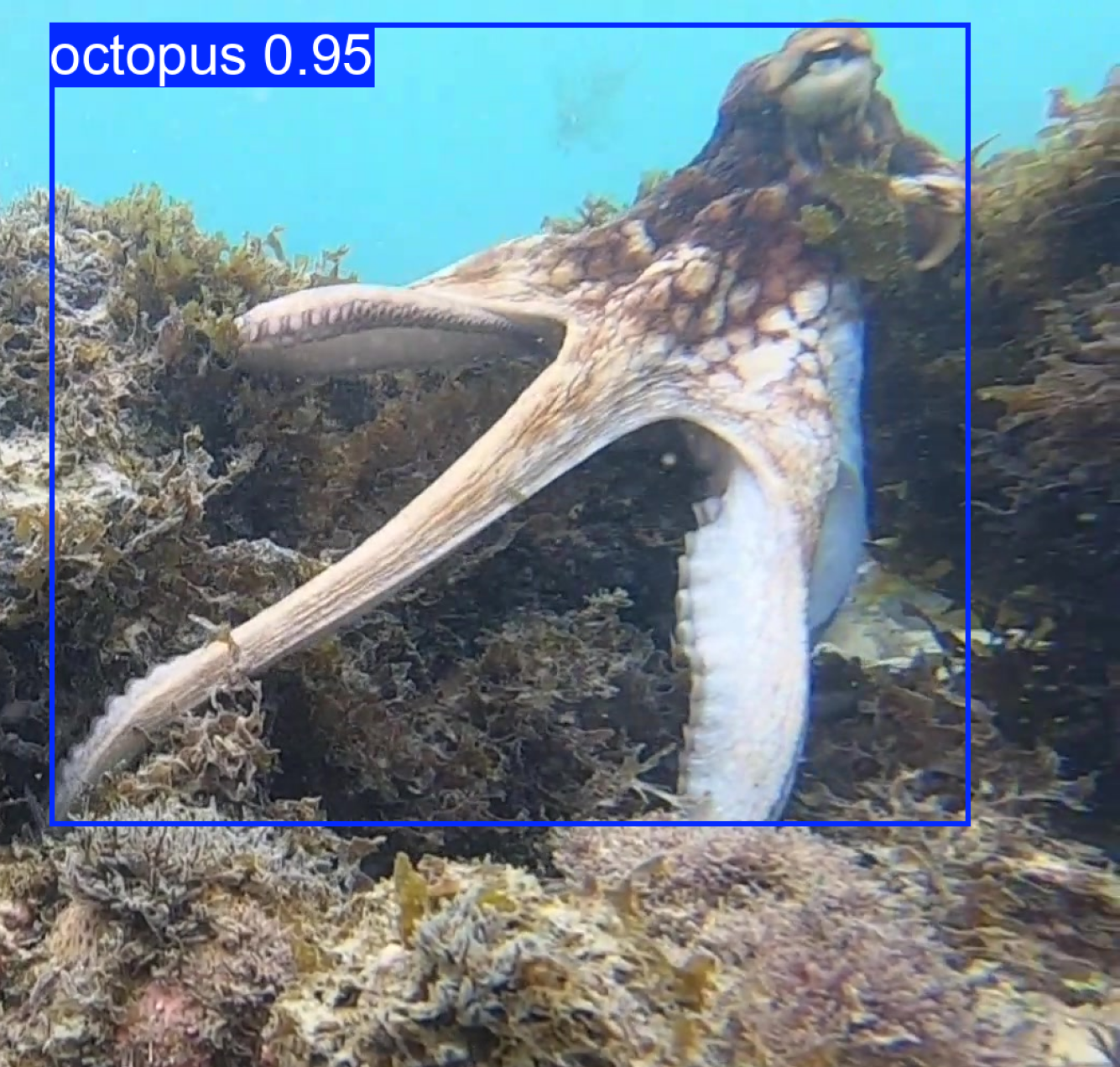}
        \caption{Frame $t+1$}
        \label{fig:yolo2}
    \end{subfigure}\hfill
    \begin{subfigure}{0.32\linewidth}
        \includegraphics[width=\linewidth]{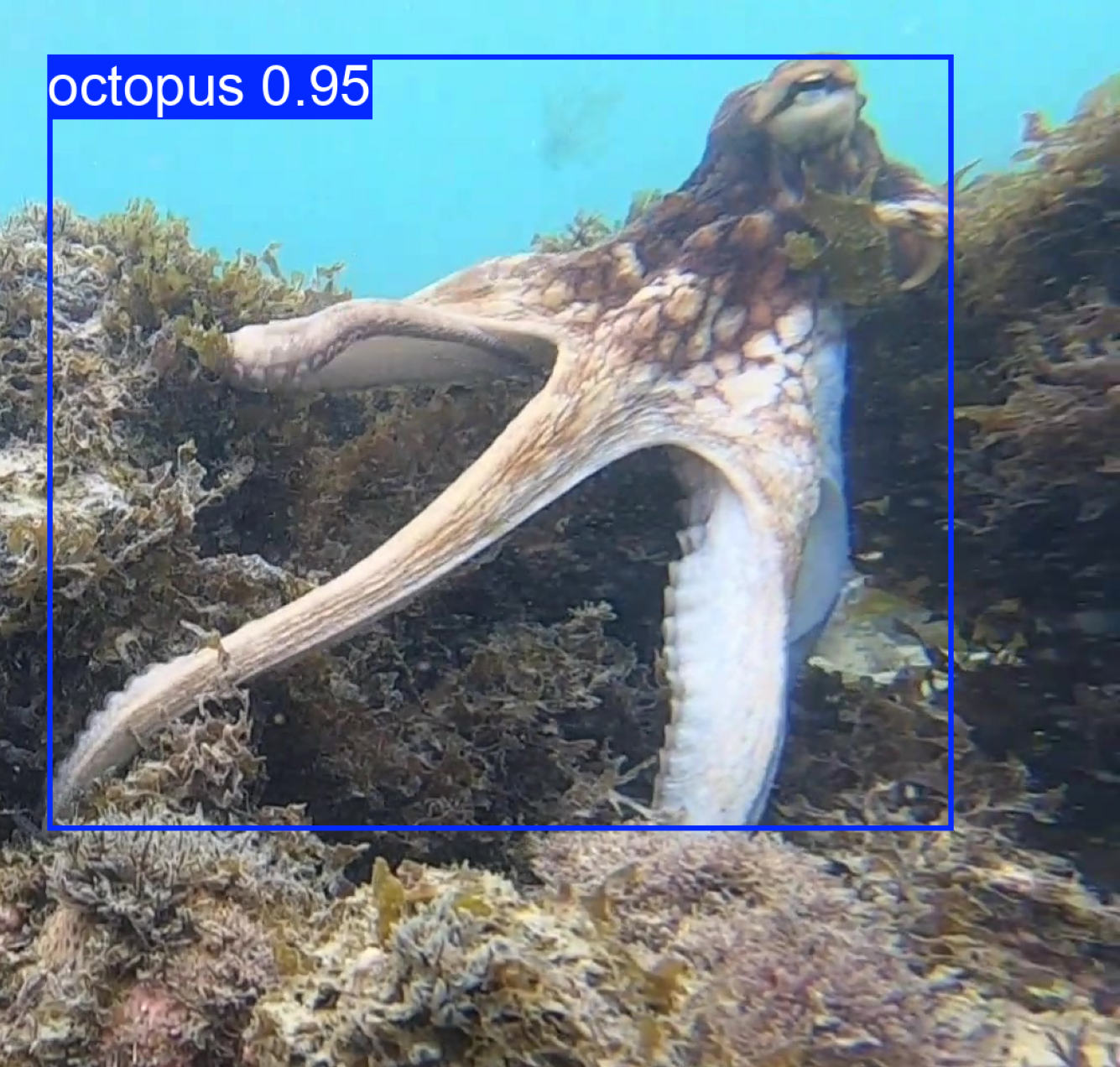}
        \caption{Frame $t+2$}
        \label{fig:yolo3}
    \end{subfigure}
    \vspace{0.5cm}  
     \setcounter{subfigure}{0} 
    \textbf{\\(C) YOLO + SAM2 output} \\
    \begin{subfigure}{0.32\linewidth}
        \includegraphics[width=\linewidth]{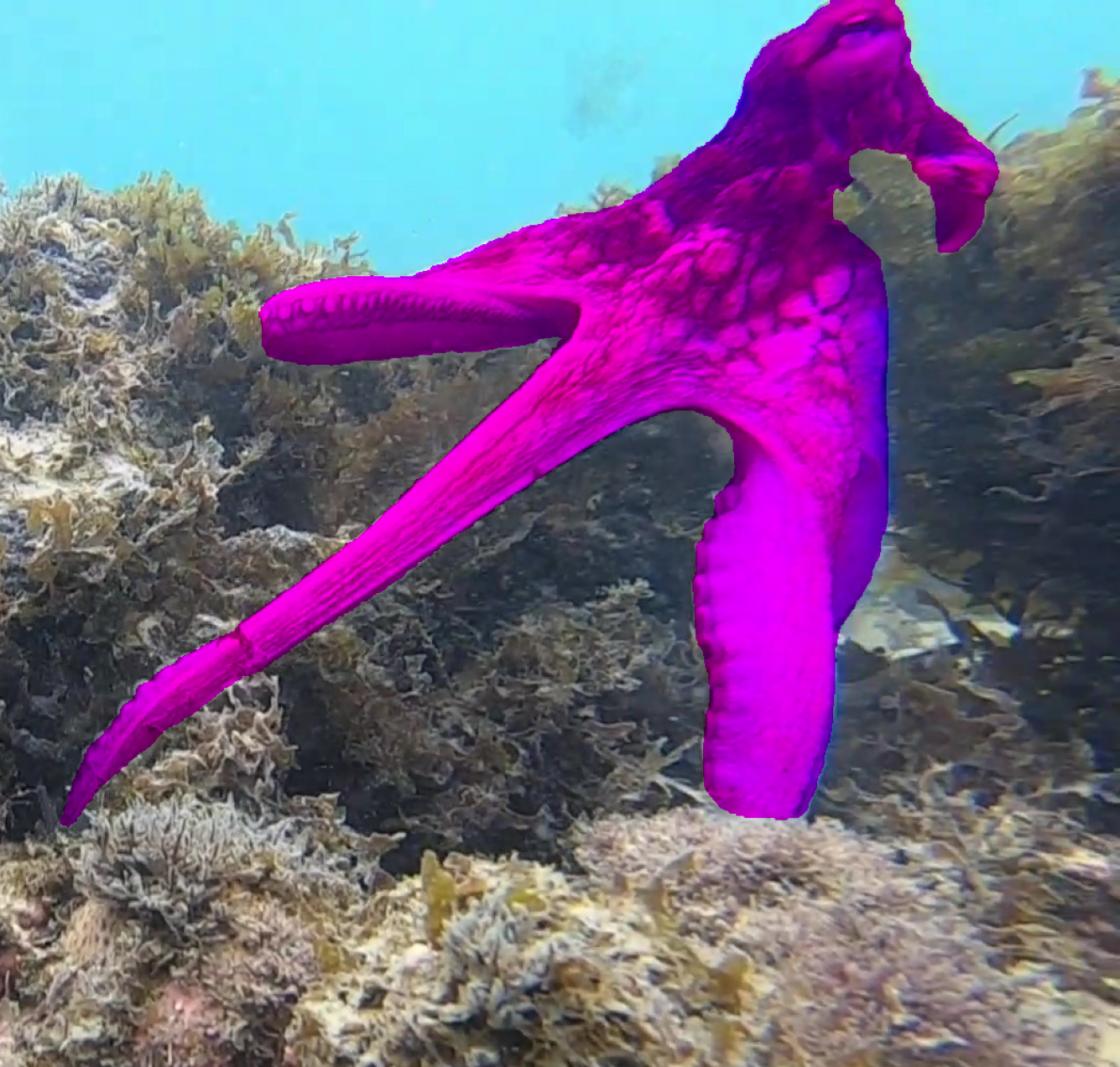}
        \caption{Frame $t$}
        \label{fig:yolo_sam1}
    \end{subfigure}\hfill
    \begin{subfigure}{0.32\linewidth}
        \includegraphics[width=\linewidth]{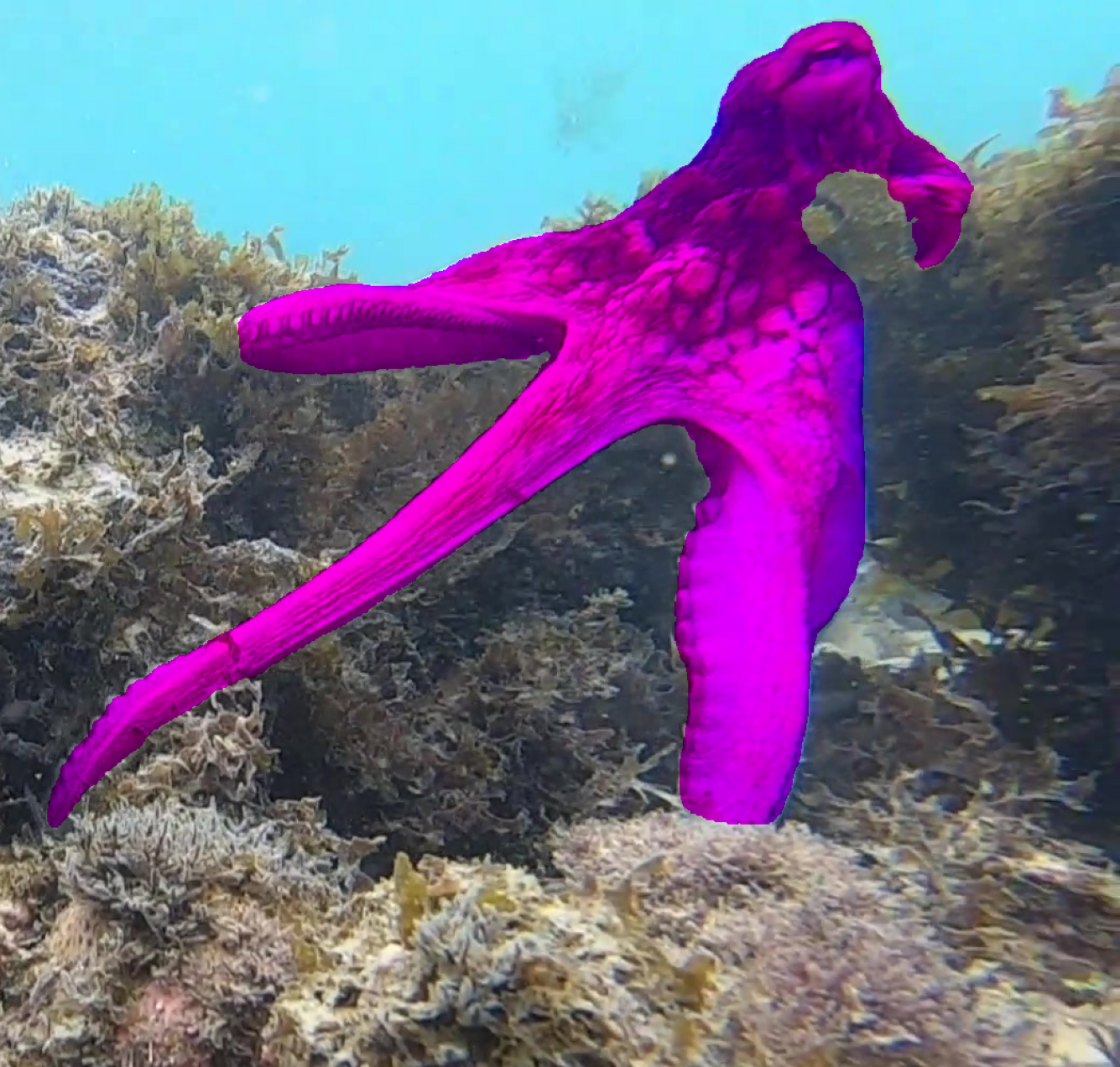}
        \caption{Frame $t+1$}
        \label{fig:yolo_sam2}
    \end{subfigure}\hfill
    \begin{subfigure}{0.32\linewidth}
        \includegraphics[width=\linewidth]{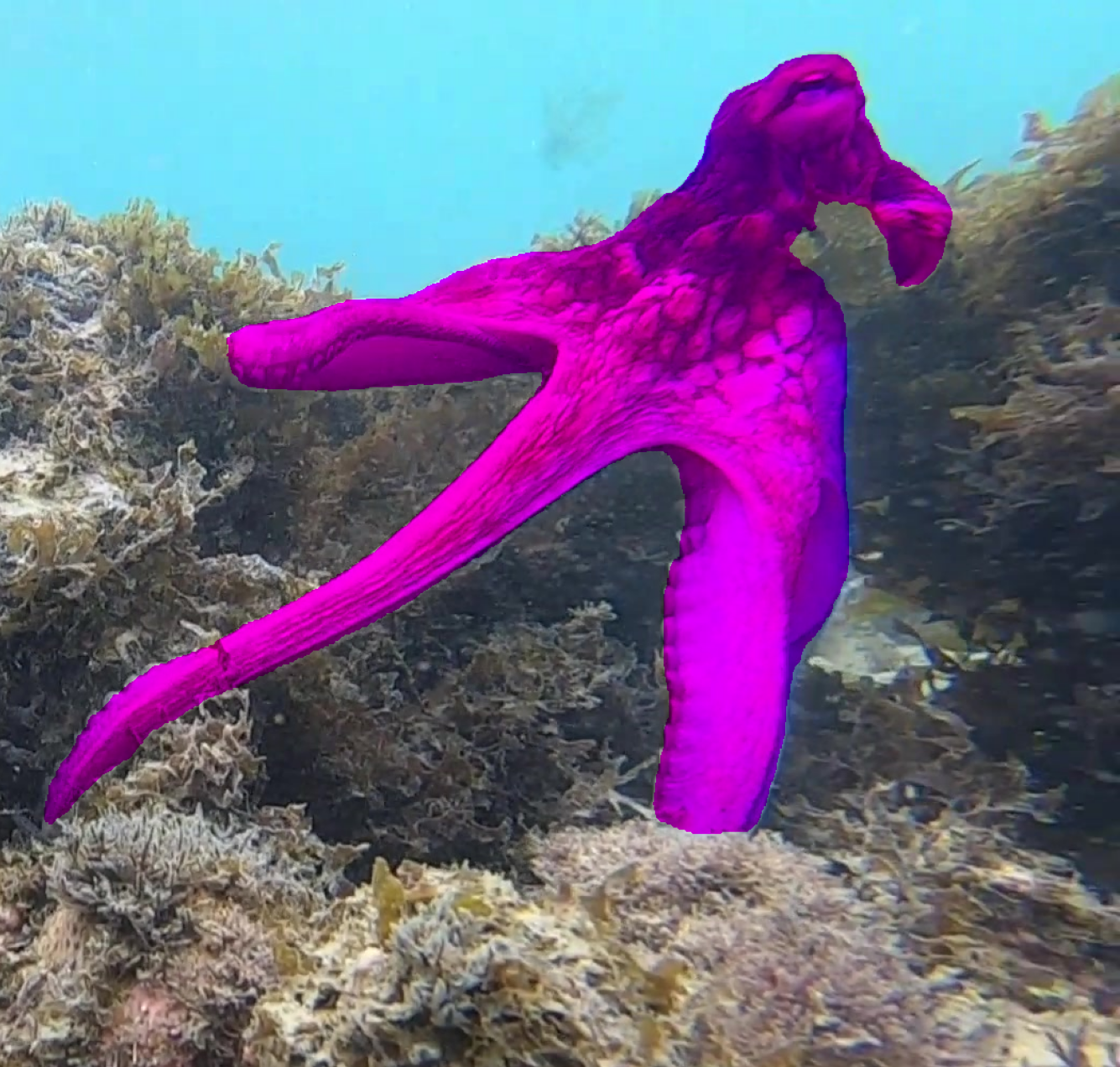}
        \caption{Frame $t+2$}
        \label{fig:yolo_sam3}
    \end{subfigure}
    \caption{Example of three consecutive frames on how to use YOLO in conjunction with SAM2. (A) SAM2 initially fails to recognize the octopus, producing a noisy, speckled segmentation mask before eventually generating a coherent result; (B) The specialized YOLO model successfully detects the octopus from the very first frame; (C) When the YOLO detections are used as prompts for SAM2, accurate segmentation masks are produced from the beginning of the sequence. Thus, we infer that the target object was not abruptly lost during processing; any degradation likely occurred gradually or along the segmentation boundaries.}
    \label{fig:3}
\end{figure}

Using the trained YOLO model, we automatically generated annotations for each frame in the videos, which can be used as prompts for SAM2, as shown in Fig.~\ref{fig:3}. However, initial tests revealed that increasing the number of annotated frames led to a rise in SAM2’s computational demands, making the method impractical for longer videos. To address this limitation, we limited the number of YOLO-annotated frames to 5, 10, or 20, depending on the experiment. These frames were uniformly sampled across the entire video, beginning with the first frame. For example, in a 100-frame video with five annotations, the selected frames would be 1, 21, 41, 61, and 81. 

\section{Results}
\subsection{Initial segmentation results}\label{sec:3.1}
Table~\ref{tab:sam2_performance} presents the results of HideAndSeg, varying the type of annotation provided. For the small SAM2 model (\texttt{SAM2.1\_hiera\_small}), the best $NC_t$ metric was achieved when both positive and negative annotations were given only on the first frame. This configuration resulted in an average of 4.05 connected components in the regions that changed between consecutive frames. A lower number reflects a lower degree of speckling in the segmentation. However, this setup also produced a high standard deviation of 16.77, indicating significant instability across the video, with some sections exhibiting sharp spikes in speckling. When we added annotations to an additional frame, we observed a deterioration in the average $NC_t$ and its standard deviation, suggesting a limitation of the model in handling increased volume or complexity of input prompts.

\begin{sidewaystable}  \caption{Performance of two SAM2 models: small (\texttt{SAM2.1\_hiera\_small}) and large (\texttt{SAM2.1\_hiera\_large}) on Different Input Types.}
  \label{tab:sam2_performance}
  \begin{tabular}{llcccc}
    \toprule
    & & \multicolumn{2}{c}{Unsupervised Metrics} & \multicolumn{2}{c}{Supervised Metrics (first frame)} \\
    \cmidrule(lr){3-4} \cmidrule(lr){5-6}
    Model & Input Type & Avg $DICE_t$ & Avg $NC_t$ & Avg $DICE$ & Avg $IoU$ \\
    \midrule
    \multirow{3}{*}{Small} & 1st frame, pos. clicks & $0.9747 \pm 0.0367$ & $10.65 \pm 30.87$ & $0.6057 \pm 0.3730$ & $0.5330 \pm 0.3737$ \\
    & 1st frame, pos. \& neg. clicks & $0.9671 \pm 0.0418$ & $4.05 \pm 16.77$ & $0.7990 \pm 0.2615$ & $0.7266 \pm 0.2862$ \\
    & + additional frame & $0.9667 \pm 0.0458$ & $5.86 \pm 52.55$ & $0.7990 \pm 0.2615$ & $0.7266 \pm 0.2862$ \\
    \midrule
    \multirow{3}{*}{Large} & 1st frame, pos. clicks & $0.9688 \pm 0.0418$ & $2.97 \pm 10.94$ & $0.9129 \pm 0.1344$ & $0.8592 \pm 0.1615$ \\
    & 1st frame, pos. \& neg. clicks & $0.9664 \pm 0.0459$ & $2.30 \pm 3.58$ & $0.9405 \pm 0.0902$ & $0.8965 \pm 0.1055$ \\
    & + additional frame & $0.9672 \pm 0.0402$ & $2.17 \pm 3.32$ & $0.9405 \pm 0.0902$ & $0.8965 \pm 0.1055$ \\
    \bottomrule
  \end{tabular}
\end{sidewaystable}

By switching to the large SAM2 model (\texttt{SAM2.1\_hiera\_large}), the segmentation quality increased. Notably, the worst-performing configuration (only positive annotations in the first frame) already outperformed the best configuration of the small model. The best performance for the large model was achieved by including annotations on one additional frame beyond the first, resulting in an average $NC_t$ of 2.17 with a standard deviation of 3.32, indicating both high segmentation quality and greater temporal stability. 

Regarding the $DICE_t$ metric, all configurations consistently yielded high scores, averaging 96\% to 97\%, which suggests that the segmentation mask area remained stable over time, regardless of the segmentation method used.

The results obtained for the supervised metrics were consistent with those observed in the unsupervised ones. Similarly, switching to the large model led to a notable improvement in segmentation consistency on the first frame. The high metric values observed in the first frame, especially with the large model, support the conclusion that SAM2 is capable of generating stable and high-quality masks throughout the entire video duration.

\subsection{Object detection results}
\begin{figure}[htbp]
    \centering
    \includegraphics[width=1\linewidth]{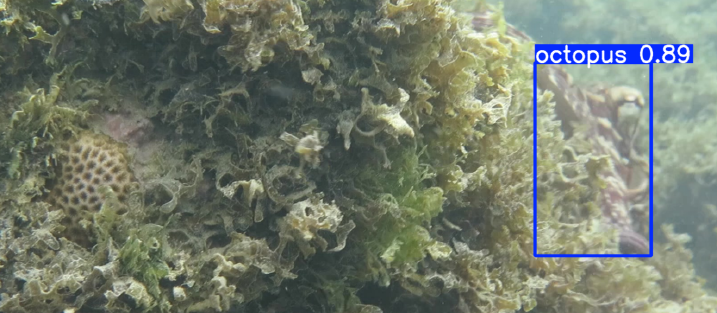}
    \caption{YOLO successfully detects the octopus on the right side of the image after being trained on a specialized dataset.}
    \label{fig:4}
\end{figure}

The YOLO object detection model, trained using segmentations generated by the best-performing configuration — SAM2 large with annotations on two frames, as described in Sec.~\ref{sec:3.1} — achieved strong results on the test set. The model demonstrated excellent performance, with a precision of 0.958, a recall of 0.968, an mAP@50 of 0.971, and an mAP@50–95 of 0.872. These results demonstrate that the model can effectively handle the challenging conditions present in the test set, including camouflage, visual obstructions, and poor water visibility. Figure~\ref{fig:4} illustrates one such scenario in which YOLO successfully detects the octopus despite the surrounding environment obscuring a significant portion of its body. 

\subsection{Fully automated segmentation results}

Table~\ref{tab:pipeline_performance} presents the segmentation results on the test set using both annotation methods from the proposed pipeline. Although the overall metrics were similar, using YOLO to prompt SAM2 improved the average $NC_t$ value. The best performance was observed when using five annotated frames. As with the results from the initial segmentation, a saturation point appears to exist when increasing the number of annotated frames, as a slight decline in performance was observed when using 10 or 20 frames. The use of YOLO also improved the supervised segmentation metrics calculated for the first frame of each video, suggesting that potential errors introduced during the manual annotation process may have been corrected by the automatic prompting method.

\begin{sidewaystable}
  \caption{Performance of the proposed pipeline on the test set.}
  \label{tab:pipeline_performance}
  \begin{tabular}{llcccc}
    \toprule
    & & \multicolumn{2}{c}{Unsupervised Metrics} & \multicolumn{2}{c}{Supervised Metrics (first frame)} \\
    \cmidrule(lr){3-4} \cmidrule(lr){5-6}
    Annotation Type & Number of annotated\\ frames  & Avg $DICE_t$ & Avg $NC_t$ & Avg $DICE$ & Avg $IoU$ \\
    \midrule
    Manual & 2 frames & $0.9695 \pm 0.0320$ & $2.75 \pm 3.74$ & $0.8997 \pm 0.2203$ & $0.8582 \pm 0.2159$ \\
    \midrule
    \multirow{3}{*}{YOLO} & 5 frames & $0.9709 \pm 0.0302$ & $2.16 \pm 2.23$ & $0.9677 \pm 0.0191$ & $0.9383 \pm 0.0349$ \\
    & 10 frames & $0.9709 \pm 0.0289$ & $2.23 \pm 2.39$ & $0.9677 \pm 0.0191$ & $0.9383 \pm 0.0349$ \\
    & 20 frames & $0.9706 \pm 0.0286$ & $2.21 \pm 2.45$ & $0.9677 \pm 0.0191$ & $0.9383 \pm 0.0349$ \\
    \bottomrule
  \end{tabular}
\end{sidewaystable}

\begin{figure}[htbp]
    \centering
    \includegraphics[width=1\linewidth]{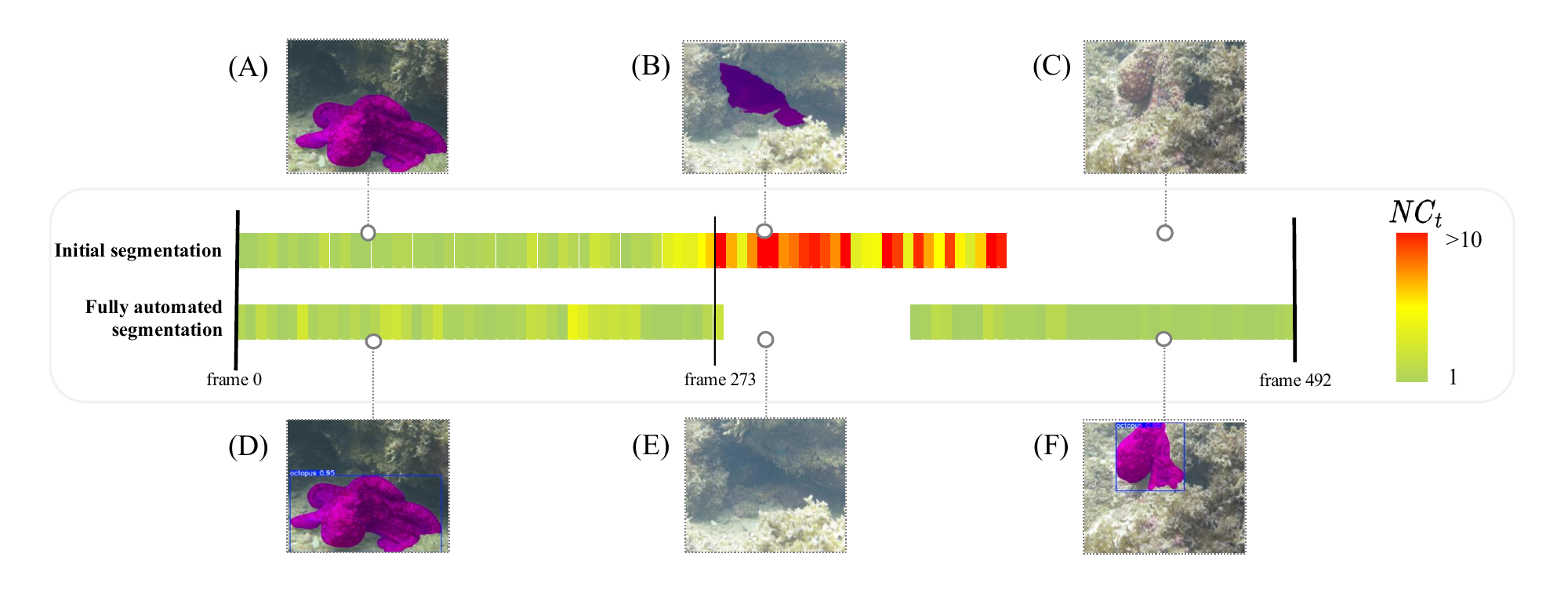}
    \caption{Variation in the $NC_t$ metric across a test video using both methods of the proposed pipeline. Initial segmentation: (A) Initially, the model produces a coherent mask, resulting in a low $NC_t$ value; (B) In the central section of the video, the octopus hides behind rocks. This behavior degrades mask quality as parts of the surrounding environment begin to be erroneously included; (C) Although the octopus becomes visible again, the noise introduced earlier prevents successful recognition, leading to an empty mask. Fully automated segmentation; (D) Initially, the method produces a coherent mask with low $NC_t$; (E) When the octopus hides, YOLO fails to detect it, resulting in an empty mask; (F) Once the octopus reappears, YOLO successfully detects it and prompts SAM2 to generate accurate masks again.}
    \label{fig:5}
\end{figure}

In addition to improving the segmentation metric, incorporating YOLO offers a significant advantage in the context of octopus segmentation. The object detection step operates independently on each frame without relying on temporal consistency. As a result, the octopus can be detected even if its shape, coloration, or texture differs significantly from previous appearances. Moreover, since the videos are recorded in natural underwater environments, occlusions, such as the octopus hiding behind rocks or other marine animals passing in front of the camera, frequently occur. In such cases, the octopus may remain out of sight for extended periods, and upon reappearing, it can look markedly different from before. This behavior was observed in one of the test set videos, with detailed results shown in Fig.~\ref{fig:5}. Using YOLO in this scenario enabled the segmentation masks to be correctly generated once the octopus re-emerged. In contrast, SAM2, relying solely on manual annotations of previous frames, failed to recover, producing empty masks even after the octopus became visible again. 

\section{Discussion}
Understanding how animals behave is crucial to conservation biology, and identifying the behaviors of wild animals has become an essential task for ecologists and conservation projects. However, quantifying the behavior of wild animals is quite challenging. This challenge arises from the substantial amount of work required to analyze the data collected in non-controlled environments~\citep{Schindler2024}. Deep neural networks have been increasingly used in animal behavior and ecology studies, as they can automatically analyze images and videos to track animals' positions, gazes, and activities, such as hunting or resting, allowing for the investigation of social behavior, interactions between individuals, and even the collective behavior of entire species~\citep{Christin2019}. Machine learning (ML) specifically for investigating marine science issues enables the solution of complex problems, processing large datasets in less time, and sometimes achieving better performance than human experts~\citep{Beyan2020}. In marine environments, ML has already been used to distinguish and count focal organisms and detect movement of sea turtles, seals, gannets, and sharks~\citep{Dujon2021,Lalgudi2025}. Since at least 80\% of animal phyla are aquatic~\citep{Hickman2014}, there is ample room for the development and refinement of computational tools that address the challenges of data analysis in marine environments.

For any automatic classification of wild octopus body patterns or behaviors, it is necessary, initially, to understand where the octopus is located in images or video frames and the surrounding environment. Reiter et al.~\citep{Reiter2018} studied cuttlefish in a laboratory setting, where the animals swam around the tank. The researchers recorded the individuals and segmented the chromatophores from the images, identifying two distinct clusters: 'dark' and 'light', which enabled the development of a tool to track chromatophores in cephalopods, allowing for studies at cellular resolution. For octopuses, two studies have already developed tools using ML, one that tracks the direction of gaze in O. bimaculoides~\citep{Taylor2020} and another that quantifies the number of O. tetricus, classifying and extracting them in images~\citep{Schneider2020}.

The present work represents the first step towards automating the detection of animal behavior in pre-recorded videos captured in natural environments. The aim was to develop a tool that would eliminate the need for lengthy manual labeling and coding, which are prone to errors and biases. While previous studies have aimed to segment octopuses in laboratories~\citep{Taylor2020}, we addressed the challenge of segmenting octopuses in their natural, uncontrolled environments. Another contribution of this work is the use of unsupervised segmentation metrics to assess the quality of the results, since there is no large publicly available ground-truth dataset for octopuses in their natural habitat. This strategy allows the methodology to be applied to new datasets without the need for extensive and costly manual annotation. HideAndSeg is more robust than the YOLO or SAM2 models alone, as it can address the specific visual challenges of underwater environments: the octopuses' ability to camouflage themselves and their non-rigid body deformations, as well as environmental issues such as turbidity and occlusion by other organisms or rocks. Using YOLO enables the model to recover from octopus detection even after prolonged periods of occlusion.

Although HideAndSeg has achieved notable results, we plan to address several limitations in future work. For example, we noted some metric limitations. The $NC_t$ metric does not account for the completeness of the segmentation. Consequently, a conservative segmentation model consistently produces well-defined but partial masks that may still yield high $NC_t$ scores, which could hide the true quality of the segmentation, particularly in cases where significant portions of the object are omitted. Another challenge will be handling the simultaneous segmentation of multiple octopuses. Since the identification model does not perform individual instance association, one option would be to annotate all octopuses and produce a union of their respective segmentation masks. Alternatively, the pipeline would require an additional step to distinguish and label individual instances, ensuring that annotations target distinct objects. 

Qualitative analysis revealed challenges in using bounding box prompts in SAM2, particularly in identifying regions of the octopus with markedly different coloration. The funnel, for example, often appears in a distinct color compared to the rest of the body, and is frequently excluded from the resulting segmentation mask. This finding suggests that bounding box prompts may be less effective in capturing the object’s fine-grained or chromatically diverse regions. 

\section{Conclusion}
This work presents HideAndSeg, an AI-based tool for video segmentation in real-life scenarios where no labeled data is available, requiring minimal manual annotation effort. To support this, we introduce unsupervised segmentation metrics that can provide meaningful insights into mask quality without relying on ground-truth annotations. 

We demonstrate the applicability of HideAndSeg to the challenging task of segmenting octopuses in underwater videos recorded in natural environments. Our approach successfully handles obstacles such as dynamic changes in position, visual obstruction, and camouflage. The results show that the annotation-free pipeline based on YOLO and SAM2 achieves segmentation performance comparable to, or in some cases, better than that of the method dependent on manual annotations. Notably, HideAndSeg performs well even in complex scenarios where the target object remains occluded or absent for a significant portion of the video. 

Future work includes applying the proposed methodology to other datasets and further specializing the octopus detector to capture specific behaviors and body patterns. Additionally, there is room for advancing the unsupervised metrics, particularly by benchmarking them against traditional segmentation metrics in datasets where ground-truth labels are available, and also for handling more than a single octopus simultaneously. As Santos \& Gois~\citep{Santos2025} note, AI tools like HideAndSeg can be catalysts for understanding biodiversity. We can use AI to manage the vast amount of data in biology. Embracing these technological advancements is crucial for addressing the enormous scope of questions biologists pose about nature and the future of our environment.

\subsection*{Acknowledgment}
This work was supported by the Wild Animal Initiative (Grant Number W-8BEN and SG24-015; MPA), Conselho Nacional de Desenvolvimento Científico e Tecnológico (CNPq), Brazil (403280/2025 and 304027/2022-7, CMDS), and Coordenação de Aperfeiçoamento de Pessoal de Nível Superior (CAPES), Brazil  Finance Code 001 (MPA, AG).

\bibliographystyle{abbrvnat}
\bibliography{hideandseg}

@InProceedings{Carion2020,
author="Carion, Nicolas
and Massa, Francisco
and Synnaeve, Gabriel
and Usunier, Nicolas
and Kirillov, Alexander
and Zagoruyko, Sergey",
editor="Vedaldi, Andrea
and Bischof, Horst
and Brox, Thomas
and Frahm, Jan-Michael",
title= {{End-to-End Object Detection with Transformers}},
booktitle="Computer Vision -- ECCV 2020",
year="2020",
publisher="Springer International Publishing",
address="Cham",
pages="213--229",
doi="10.1007/978-3-030-58452-8\_13",
isbn="978-3-030-58452-8"
}

@article{Yu2025,
  author    = {Yu, Minghui and Xie, Zhijun and Ye, Yangfang and Shi, Ce},
  year      = {2025},
  month     = {jul},
  title     = {{TMAE-YOLO: precision detection of mud crabs in underwater environments}},
  journal   = {Aquaculture International},
  pages     = {462},
  volume    = {33},
  number    = {6},
  issn      = {1573-143X},
  url       = {https://doi.org/10.1007/s10499-025-02149-8},
  doi       = {10.1007/s10499-025-02149-8}
}

@article{Andrade2023,
AUTHOR = {Andrade, Michaella P. and Santos, Charles Morphy D. and De Paiva, Mizziara M. M. and Medeiros, Sylvia L. S. and O’Brien, C. E. and Lima, Françoise D. and Machado, Janaina F. and Leite, Tatiana S.},
TITLE = {{Assessing Negative Welfare Measures for Wild Invertebrates: The Case for Octopuses}},
JOURNAL = {Animals},
VOLUME = {13},
YEAR = {2023},
NUMBER = {19},
pages = {3021},
URL = {https://www.mdpi.com/2076-2615/13/19/3021},
PubMedID = {37835627},
ISSN = {2076-2615},
DOI = {10.3390/ani13193021}
}

@article{Beyan2020,
    author = {Beyan, Cigdem and Browman, Howard I},
    title = {{Setting the stage for the machine intelligence era in marine science}},
    journal = {ICES Journal of Marine Science},
    volume = {77},
    number = {4},
    pages = {1267-1273},
    year = {2020},
    month = {06},
    issn = {1054-3139},
    doi = {10.1093/icesjms/fsaa084},
    url = {https://doi.org/10.1093/icesjms/fsaa084},
    eprint = {https://academic.oup.com/icesjms/article-pdf/77/4/1267/33513046/fsaa084.pdf},
}

@article{Chen2024,
title = {{YOLO-SAG: An improved wildlife object detection algorithm based on YOLOv8n}},
journal = {Ecological Informatics},
volume = {83},
pages = {102791},
year = {2024},
issn = {1574-9541},
doi = {https://doi.org/10.1016/j.ecoinf.2024.102791},
url = {https://www.sciencedirect.com/science/article/pii/S1574954124003339},
author = {Lingli Chen and Gang Li and Shunkai Zhang and Wenjie Mao and Mei Zhang}
}

@article{Christin2019,
author = {Christin, Sylvain and Hervet, \'Eric and Lecomte, Nicolas},
title = {{Applications for deep learning in ecology}},
journal = {Methods in Ecology and Evolution},
volume = {10},
number = {10},
pages = {1632-1644},
doi = {https://doi.org/10.1111/2041-210X.13256},
url = {https://besjournals.onlinelibrary.wiley.com/doi/abs/10.1111/2041-210X.13256},
eprint = {https://besjournals.onlinelibrary.wiley.com/doi/pdf/10.1111/2041-210X.13256},
year = {2019}
}

@article{Cortes2025,
    author = {Côrtes, Mariana Osório and Santos, Bruna Baleeiro dos and Freitas, Renato Hajenius Aché de and O’Brien, C E and Leite, Tatiana Silva},
    title = {{Review of the ecological aspects of sympatric species Octopus americanus and Octopus insularis (Cephalopoda: Octopodidae) in the Western Atlantic}},
    journal = {Current Zoology},
    pages = {zoaf023},
    year = {2025},
    month = {05},
    issn = {2396-9814},
    doi = {10.1093/cz/zoaf023},
    url = {https://doi.org/10.1093/cz/zoaf023},
    eprint = {https://academic.oup.com/cz/advance-article-pdf/doi/10.1093/cz/zoaf023/63215720/zoaf023.pdf},
}

@article{Dujon2021,
author = {Dujon, Antoine M. and Ierodiaconou, Daniel and Geeson, Johanna J. and Arnould, John P. Y. and Allan, Blake M. and Katselidis, Kostas A. and Schofield, Gail},
title = {{Machine learning to detect marine animals in UAV imagery: effect of morphology, spacing, behaviour and habitat}},
journal = {Remote Sensing in Ecology and Conservation},
volume = {7},
number = {3},
pages = {341-354},
doi = {https://doi.org/10.1002/rse2.205},
url = {https://zslpublications.onlinelibrary.wiley.com/doi/abs/10.1002/rse2.205},
eprint = {https://zslpublications.onlinelibrary.wiley.com/doi/pdf/10.1002/rse2.205},
year = {2021}
}

@article{Flash2023,
    author = {Flash, Tamar and Zullo, Letizia},
    title = {{Biomechanics, motor control and dynamic models of the soft limbs of the octopus and other cephalopods}},
    journal = {Journal of Experimental Biology},
    volume = {226},
    number = {Suppl\_1},
    pages = {jeb245295},
    year = {2023},
    month = {04},
    issn = {0022-0949},
    doi = {10.1242/jeb.245295},
    url = {https://doi.org/10.1242/jeb.245295},
    eprint = {https://journals.biologists.com/jeb/article-pdf/226/Suppl\_1/jeb245295/2793398/jeb245295.pdf},
}

@book{Hickman2014,
  author    = {Cleveland Hickman and Larry Roberts and Susan Keen and Allan Larson and David Eisenhour},
  title     = {{Animal Diversity}},
  year      = {2014},
  publisher = {McGraw-Hill},
  address   = {NY},
  pages    = { 512},
  edition = {7th}
}

@Inbook{Ikeda2021,
author="Ikeda, Yuzuru",
editor="Hashimoto, Hisashi
and Goda, Makoto
and Futahashi, Ryo
and Kelsh, Robert
and Akiyama, Toyoko",
title= {{Color Change in Cephalopods}},
bookTitle= {{Pigments, Pigment Cells and Pigment Patterns}},
year="2021",
publisher="Springer Singapore",
address="Singapore",
pages="425--449",
isbn="978-981-16-1490-3",
doi="10.1007/978-981-16-1490-3\_14",
url="https://doi.org/10.1007/978-981-16-1490-3\_14"
}

@misc{Khanam2024,
      title={{YOLOv11: An Overview of the Key Architectural Enhancements}}, 
      author={Rahima Khanam and Muhammad Hussain},
      year={2024},
      eprint={2410.17725},
      archivePrefix={arXiv},
      primaryClass={cs.CV},
      url={https://arxiv.org/abs/2410.17725}, 
}

@misc{Kirillov2023,
      title={{Segment Anything}}, 
      author={Alexander Kirillov and Eric Mintun and Nikhila Ravi and Hanzi Mao and Chloe Rolland and Laura Gustafson and Tete Xiao and Spencer Whitehead and Alexander C. Berg and Wan-Yen Lo and Piotr Dollár and Ross Girshick},
      year={2023},
      eprint={2304.02643},
      archivePrefix={arXiv},
      primaryClass={cs.CV},
      url={https://arxiv.org/abs/2304.02643}, 
}

@article{Lalgudi2025,
author = {Lalgudi, Chinmay K. and Leone, Mark E. and Clark, Jaden V. and Madrigal-Mora, Sergio and Espinoza, Mario},
title = {{Zero-shot shark tracking and biometrics from aerial imagery}},
journal = {Methods in Ecology and Evolution},
volume = {16},
number = {9},
pages = {2023-2035},
doi = {https://doi.org/10.1111/2041-210X.70116},
url = {https://besjournals.onlinelibrary.wiley.com/doi/abs/10.1111/2041-210X.70116},
eprint = {https://besjournals.onlinelibrary.wiley.com/doi/pdf/10.1111/2041-210X.70116},
year = {2025}
}

@article{Messenger2001,
author = {Messenger, John. B.},
title = {{Cephalopod chromatophores: neurobiology and natural history}},
journal = {Biological Reviews},
volume = {76},
number = {4},
pages = {473-528},
doi = {https://doi.org/10.1017/S1464793101005772},
url = {https://onlinelibrary.wiley.com/doi/abs/10.1017/S1464793101005772},
eprint = {https://onlinelibrary.wiley.com/doi/pdf/10.1017/S1464793101005772},
year = {2001}
}

@article{OBrien2021,
place={Auckland, New Zealand}, 
title={{A field guide to distinguishing Octopus insularis and Octopus americanus (Octopoda: Octopodidae)}}, 
volume={5060}, 
url={https://mapress.com/zt/article/view/zootaxa.5060.4.8}, 
DOI={10.11646/zootaxa.5060.4.8}, 
number={4}, 
journal={Zootaxa}, 
author={O’brien, Caitlin E. and Bennice, Chelsea O. and Leite, Tatiana}, 
year={2021}, 
month={Nov.}, 
pages={589–594} 
}

@misc{Ravi2024,
     title={{SAM 2: Segment Anything in Images and Videos}}, 
      author={Nikhila Ravi and Valentin Gabeur and Yuan-Ting Hu and Ronghang Hu and Chaitanya Ryali and Tengyu Ma and Haitham Khedr and Roman Rädle and Chloe Rolland and Laura Gustafson and Eric Mintun and Junting Pan and Kalyan Vasudev Alwala and Nicolas Carion and Chao-Yuan Wu and Ross Girshick and Piotr Dollár and Christoph Feichtenhofer},
      year={2024},
      eprint={2408.00714},
      archivePrefix={arXiv},
      primaryClass={cs.CV},
      url={https://arxiv.org/abs/2408.00714}, 
}

@INPROCEEDINGS{Redmon2015,
author = { Redmon, Joseph and Divvala, Santosh and Girshick, Ross and Farhadi, Ali },
booktitle = {2016 IEEE Conference on Computer Vision and Pattern Recognition (CVPR) },
title = {{You Only Look Once: Unified, Real-Time Object Detection}},
year = {2016},
volume = {},
ISSN = {1063-6919},
pages = {779-788},
doi = {10.1109/CVPR.2016.91},
url = {https://doi.ieeecomputersociety.org/10.1109/CVPR.2016.91},
publisher = {IEEE Computer Society},
address = {Los Alamitos, CA, USA},
month =Jun}

@article{Reiter2018,
  author    = {Reiter, Sam and Hülsdunk, Philipp and Woo, Theodosia and Lauterbach, Marcel A. and Eberle, Jessica S. and Akay, Leyla Anne and Longo, Amber and Meier-Credo, Jakob and Kretschmer, Friedrich and Langer, Julian D. and Kaschube, Matthias and  Laurent, Gilles},
  title     = {{Elucidating the control and development of skin patterning in cuttlefish}},
  journal   = {Nature},
  year      = {2018},
  volume    = {562},
  number    = {7727},
  pages     = {361--366},
  doi       = {10.1038/s41586-018-0591-3}
}

@article{Roy2023,
title = {{WilDect-YOLO: An efficient and robust computer vision-based accurate object localization model for automated endangered wildlife detection}},
journal = {Ecological Informatics},
volume = {75},
pages = {101919},
year = {2023},
issn = {1574-9541},
doi = {https://doi.org/10.1016/j.ecoinf.2022.101919},
url = {https://www.sciencedirect.com/science/article/pii/S1574954122003697},
author = {Arunabha M. Roy and Jayabrata Bhaduri and Teerath Kumar and Kislay Raj}
}

@article{Santos2025,
  author    = {Charles Morphy D. Santos  and João Paulo Gois},
  title     = {{Artificial Intelligence as catalyst for biodiversity understanding}},
  journal   = {Communications of the ACM},
  year      = {2025},
  volume    = {68},
  number    = {3},
  pages     = {27--29},
  doi       = {10.1145/3701570}
}

@phdthesis{Schneider2020,
  author    = {Stefan Schneider},
  title     = {{Deep Learning Based Computer Vision for Animal Re-Identification}},
  school    = {University of Guelph},
  year      = {2020},
  url       = {https://atrium.lib.uoguelph.ca/items/814db23b-c6a0-48c7-91fb-1ec578c09bd8}
}

@article{Schindler2024,
AUTHOR = {Schindler, Frank and Steinhage, Volker and van Beeck Calkoen, Suzanne T. S. and Heurich, Marco},
TITLE = {{Action Detection for Wildlife Monitoring with Camera Traps Based on Segmentation with Filtering of Tracklets (SWIFT) and Mask-Guided Action Recognition (MAROON)}},
JOURNAL = {Applied Sciences},
VOLUME = {14},
YEAR = {2024},
NUMBER = {2},
ARTICLE-NUMBER = {514},
URL = {https://www.mdpi.com/2076-3417/14/2/514},
ISSN = {2076-3417},
DOI = {10.3390/app14020514}
}

@article{Schnell2021,
author = {Schnell, Alexandra K. and Amodio, Piero and Boeckle, Markus and Clayton, Nicola S.},
title = {{How intelligent is a cephalopod? Lessons from comparative cognition}},
journal = {Biological Reviews},
volume = {96},
number = {1},
pages = {162-178},
doi = {https://doi.org/10.1111/brv.12651},
url = {https://onlinelibrary.wiley.com/doi/abs/10.1111/brv.12651},
eprint = {https://onlinelibrary.wiley.com/doi/pdf/10.1111/brv.12651},
year = {2021}
}

@article{Shook2024,
title = {{Dynamic skin behaviors in cephalopods}},
journal = {Current Opinion in Neurobiology},
volume = {86},
pages = {102876},
year = {2024},
issn = {0959-4388},
doi = {https://doi.org/10.1016/j.conb.2024.102876},
url = {https://www.sciencedirect.com/science/article/pii/S0959438824000382},
author = {Erica N. Shook and George Thomas Barlow and Daniella Garcia-Rosales and Connor J. Gibbons and Tessa G. Montague}
}

@misc{Taylor2020,
  author    = {Mark Andrew Taylor},
  title     = {{Autonomous eye tracking in Octopus bimaculoides}},
  year      = {2020},
  howpublished = {Senior Thesis, Dartmouth College},
  url       = {https://digitalcommons.dartmouth.edu/senior\_theses/151/}
}

@article{Thomas2025,
  author    = {Shiney Thomas and Elsa George and Alphonsa Francis and Anna Job and Ann Maria James},
  title     = {{Wildlife detection and recognition using YOLO V8}},
  journal   = {International Journal on Emerging Research Areas},
  year      = {2025},
  volume    = {4},
  number    = {2},
  pages     = {81--87},
  doi       = {10.5281/zenodo.14714518}
}

@article{VanderWalt2014,
  author    = {Stéfan van der Walt and Johannes L. Schönberger and Juan Nunez-Iglesias and François Boulogne and Joshua D. Warner and Neil Yager and Emmanuelle Gouillart and Tony Yu and the~scikit-image~contributors},
  title     = {{scikit-image: image processing in Python}},
  journal   = {PeerJ},
  year      = {2014},
  volume    = {2},
  pages     = {e453},
  doi       = {10.7717/peerj.453}
}

@article{Vijayalakshmi2024,
  author    = {Vijayalakshmi Mohankumar, Sasithradevi Anbalagan},
  title     = {{A comprehensive review on deep learning architecture for pre-processing of underwater images}},
  journal   = {SN Computer Science},
  year      = {2024},
  volume    = {5},
  number    = {472},
  doi       = {10.1007/s42979-024-02847-9}
}

@article{Zheng2024,
title = {{A video object segmentation-based fish individual recognition method for underwater complex environments}},
journal = {Ecological Informatics},
volume = {82},
pages = {102689},
year = {2024},
issn = {1574-9541},
doi = {https://doi.org/10.1016/j.ecoinf.2024.102689},
url = {https://www.sciencedirect.com/science/article/pii/S1574954124002310},
author = {Tao Zheng and Junfeng Wu and Han Kong and Haiyan Zhao and Boyu Qu and Liang Liu and Hong Yu and Chunyu Zhou}
}

\end{document}